\documentclass{article}
\usepackage{ijcai18}

\usepackage{times}
\usepackage{xcolor}
\usepackage{soul}
\usepackage{url}
\usepackage[hidelinks]{hyperref}
\usepackage[utf8]{inputenc}
\usepackage[small]{caption}
\usepackage{graphicx}
\usepackage{amsmath}
\usepackage{amsfonts}
\usepackage{amssymb}
\usepackage{subcaption}
\usepackage{tabularx}
\usepackage[bottom]{footmisc}

\title{ElimiNet: A Model for Eliminating Options for Reading Comprehension with Multiple Choice Questions}

\author{
Soham Parikh\thanks{denotes equal contribution},
Ananya B.\ Sai\footnotemark[1],
Preksha Nema\footnotemark[1],
Mitesh M.\ Khapra
\\ 
Indian Institute of Technology, Madras\\
\{sohamp,ananyasb,preksha,miteshk\}@cse.iitm.ac.in}

\begin{document}

\maketitle

\begin{abstract}
  The task of Reading Comprehension with Multiple Choice Questions, requires a human (or machine) to read a given \{\textit{passage, question}\} pair and select one of the $n$ given options. The current state of the art model for this task first computes a question-aware representation for the passage and then \textit{selects} the option which has the maximum similarity with this representation. However, when humans perform this task they do not just focus on option selection but use a combination of \textit{elimination} and \textit{selection}. Specifically, a human would first try to eliminate the most irrelevant option and then read the passage again in the light of this new information (and perhaps ignore portions corresponding to the eliminated option). This process could be repeated multiple times till the reader is finally ready to select the correct option. We propose \textit{ElimiNet}, a neural network-based model which tries to mimic this process. Specifically, it has gates which decide whether an option can be eliminated given the \{\textit{passage, question}\} pair and if so it tries to make the passage representation orthogonal to this eliminated option (akin to ignoring portions of the passage corresponding to the eliminated option). The model makes multiple rounds of partial elimination to refine the passage representation and finally uses a selection module to pick the best option. We evaluate our model on the recently released large scale RACE dataset and show that it outperforms the current state of the art model on $7$ out of the $13$ question types in this dataset. Further, we show that taking an ensemble of our \textit{elimination-selection} based method with a \textit{selection} based method gives us an improvement of $3.1\%$ over the best-reported performance on this dataset.
\end{abstract}

\section{Introduction}
Reading comprehension is the task of answering questions pertaining to a given passage. An AI agent which can display such capabilities would be useful in a wide variety of commercial applications such as answering questions from financial reports of a company, troubleshooting using product manuals, answering general knowledge questions from Wikipedia documents, \textit{etc}. Given its widespread applicability, several variants of this task have been studied in the literature. For example, given a passage and a question, the answer could either (i) match some span in the passage or (ii) be generated from the passage or (iii) be one of the $n$ given candidate answers. The last variant is typically used in various high school, middle school, and competitive examinations. We refer to this as Reading Comprehension with Multiple Choice Questions (RC-MCQ). There is an increasing interest in building AI agents with deep language understanding capabilities which can perform at par with humans on such competitive tests. For example, recently \cite{DBLP:conf/emnlp/LaiXLYH17} have released a large scale dataset for RC-MCQ collected from high school and middle school English examinations in China comprising of approximately $28000$ passages and $100000$ questions. The large size of this dataset makes it possible to train and evaluate complex neural network based models and measure the scientific progress on RC-MCQ.
\begin{figure}
\fbox{\begin{minipage}{23em}\small
Passage: \color{red}{One day, I was studying at home. Suddenly, there was a loud noise}...\color{black}{A building in my neighborhood was on fire...A few people jumped out of the window... Those who were still on the second floor were just crying for help...Firefighters arrived at last. They fought the fire bravely.} \color{blue}{Water pipes were used and a ladder was put near the second-floor window.}\color{green}{ Then the people inside were taken out by the firefighters}\color{black}{...}\color{orange}{Thanks to the firefighters, the people inside were saved and the fire was put out in the end}\color{black}{, but many things, such as desk, pictures and clothes, were damaged.} \newline
\color{black}{
\textit{\textbf{Question:} How did the people who didn't jump out of the window get out of the building?}} \newline \color{black}{\textbf{Option A:}} \color{green}{They were taken out by the firefighters.}\\ \color{black}{\textbf{Option B:}} \color{blue}{They climbed down a ladder by themselves.}\\ \color{black}{\textbf{Option C:}} \color{orange}{They walked out after the fire was put out.}\\ \color{black}{\textbf{Option D:} They were taken out by doctors}\\ \textbf{Correct Option:} A
\end{minipage}}
\caption{Example of RC-MCQ from RACE dataset}
\label{fig:example}
\end{figure}
While answering such Multiple Choice Questions (MCQs) (\textit{e.g.}, Figure \ref{fig:example}), humans typically use a combination of \textit{option elimination} and  \textit{option selection}. More specifically, it makes sense to first try to eliminate options which are completely irrelevant to the given question. While doing so, we may also be able to discard certain portions of the passage which are not relevant to the question (because they revolve around the option which has been eliminated, \textit{e.g.}, portions marked in blue and orange, corresponding to Option B and Option C respectively in Figure ~\ref{fig:example}). This process can then be repeated multiple times, each time eliminating an option and refining the passage (by discarding irrelevant portions). Finally, when it is no longer possible to eliminate any option, we can pick the best option from the remaining options. In contrast, the current state of the art models for RC-MCQ focus explicitly on option selection. Specifically, given a question and a passage, they first compute a question aware representation of the passage (say $d_q$). They then compute a representation for each of the $n$ options and select an option whose representation is closest to $d_q$. There is no iterative process where options get eliminated and the representation of the passage gets refined in the light of this elimination.

We propose a model which tries to mimic the human process of answering MCQs. Similar to the existing state of the art method \cite{DBLP:conf/acl/DhingraLYCS17}, we first compute a question-aware representation of the passage (which essentially tries to retain portions of the passage which are only relevant to the question). We then use an elimination gate (depending on the passage, question and option) which takes a soft decision as to whether an option needs to be eliminated or not. Next, akin to the human process described above, we would like to discard portions of the passage representation which are aligned with this eliminated option. We do this by subtracting the component of the passage representation along the option representation (similar to Gram-Schmidt orthogonalization). The amount of orthogonalization depends on the soft decision given by the elimination gate. We repeat this process multiple times, during each pass doing a soft elimination of the options and refining the passage representation. At the end of a few passes, we expect the passage representation to be orthogonal (hence dissimilar) to the irrelevant options. Finally, we use a selection module to select the option which is most similar to the refined passage representation. We refer to this model as \textit{ElimiNet}. Note that such a model will not make sense in cases where the options are highly related. For example, if the question is about life stages of a butterfly and the options are four different orderings of the words \textit{butterfly, egg, pupa, caterpillar} then it does not make sense to orthogonalize the passage representation to the incorrect option representations. However, the dataset that we focus on in this work does not contain questions which have such permuted options.

We evaluate \textit{ElimiNet} on the RACE dataset and compare it with Gated Attention Reader (GAR) \cite{DBLP:conf/acl/DhingraLYCS17}, the current state of the art model on this dataset. We show that of the $13$ question types in this dataset our model outperforms GAR on 7 question types. We also visualize the soft elimination probabilities learnt by \textit{ElimiNet} and observe that it indeed learns to iteratively refine the passage representation and push the probability mass towards the correct option. Finally, we show that an ensemble model combining \textit{ElimiNet} with \textit{GAR} gives an accuracy of $47.2\%$ which is $3.1\%$ (relative) better than the best-reported performance on this dataset.
The code for our model is publicly available\footnote{\url{https://github.com/sohamparikh94/ElimiNet}}.

\section{Related Work}

Over the last few years, the availability of large scale datasets has led to an increasing interest in the task of Reading Comprehension. These datasets cover different variations of the Reading comprehension task. For example, SQuAD \cite{DBLP:conf/emnlp/RajpurkarZLL16}, TriviaQA \cite{DBLP:conf/acl/JoshiCWZ17}, NewsQA \cite{DBLP:journals/corr/TrischlerWYHSBS16}, MS MARCO \cite{ms-marco-human-generated-machine-reading-comprehension-dataset}, NarrativeQA \cite{DBLP:journals/corr/abs-1712-07040}, \textit{etc.} contain \{\textit{passage, question, answer}\} where the answer matches a span of the passage or it has to be generated. On the other hand, CNN/Daily Mail \cite{DBLP:conf/nips/HermannKGEKSB15}, Children's Book Test (CBT) \cite{DBLP:journals/corr/HillBCW15} and Who Did What (WDW) dataset \cite{DBLP:conf/emnlp/OnishiWBGM16} offer cloze-style RC where the task is to predict a missing word/entity (from the passage) in the question. Some other datasets such as MCTest \cite{DBLP:conf/emnlp/RichardsonBR13}, AI2 \cite{DBLP:conf/ijcai/KhashabiKSCER16} and RACE contain RC with multiple choice questions (RC-MCQ) where the task is to select the right answer.

The advent of these datasets and the general success of deep learning for various NLP tasks, has led to a proliferation of neural network based models for RC. For example, the models proposed in \cite{DBLP:journals/corr/XiongZS16,DBLP:journals/corr/SeoKFH16,DBLP:conf/acl/WangYWCZ17,DBLP:journals/corr/HuPQ17} address the first variant of RC requiring span prediction as in the SQuAD dataset. Similarly, the models proposed in \cite{DBLP:conf/acl/ChenBM16,DBLP:conf/acl/KadlecSBK16,DBLP:conf/acl/CuiCWWLH17,DBLP:conf/acl/DhingraLYCS17} address the second variant of RC requiring cloze-style QA. Finally, \cite{DBLP:conf/emnlp/LaiXLYH17} adapt the the models proposed in  \cite{DBLP:conf/acl/ChenBM16,DBLP:conf/acl/DhingraLYCS17} for cloze-style RC and use them to address the problem of RC-MCQ. Irrespective of which of the three variants of RC they address, these models use a very similar framework. Specifically, these models contain components for (i) encoding the passage (ii) encoding the question (iii) capturing interactions between the question and the passage (iv) capturing interactions between question and the options (for MCQ) (v) making multiple passes over the passage and (vi) a decoder to predict/generate/select an answer. The differences between the models arise from the specific choice of the encoder, decoder, interaction functions and iteration mechanism. Most of the current state of the art models can be seen as special instantiations of the above framework. 

The key difference between our model and existing models for RC-MCQ is that we introduce components for (soft-)eliminating irrelevant options and refining the passage representation in the light of this elimination. The passage representation thus refined over multiple (soft-)elimination rounds is then used for selecting the most relevant option. To the best of our knowledge, this is the first model which introduces the idea of option elimination for RC-MCQ.

\begin{figure}
	\includegraphics[width=0.49\textwidth]{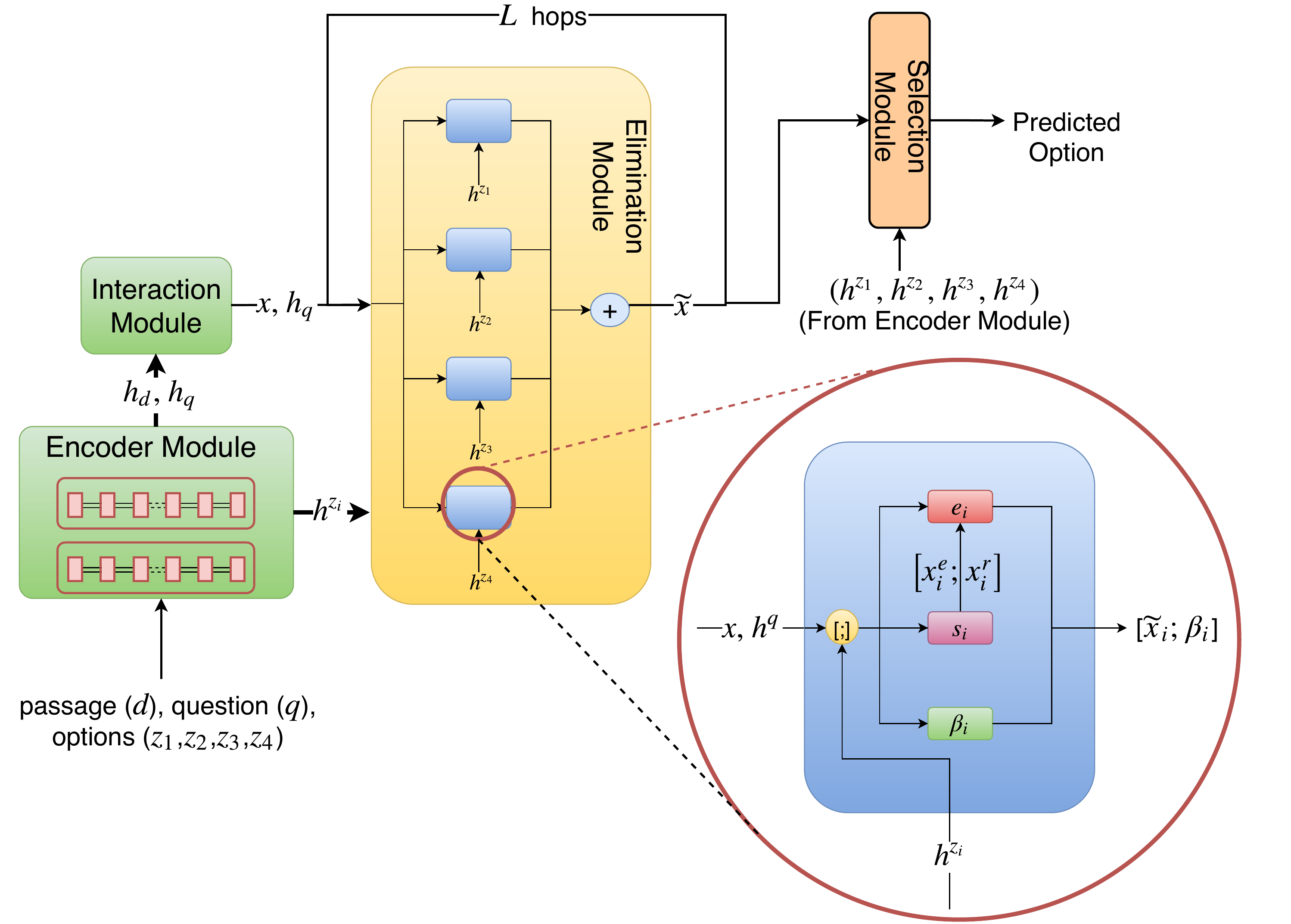}
    \caption{A simplistic diagram of the proposed model}
    \label{fig:diagram}
\end{figure}
\section{Proposed Model}
Given a passage $D = [w_1^d, w_2^d,\ldots, w_M^d]$ of word-length $M$, a question $Q = [w_1^q, w_2^q,\ldots, w_N^q ]$ of word-length $N$ and $n$ options $Z_k = [w_1^z, w_2^z, \ldots, w_{J_k}^z]$ where $1 \leqslant k \leqslant n$ and each option is of word-length $J_k$, the task is to predict a conditional probability distribution over the options (\textit{i.e.}, to predict $P(Z_i| D, Q)$). We model this distribution using a neural network which contains modules for encoding the passage/question/options, capturing the interactions between them, eliminating options and finally selecting the correct option. We refer to these as the encoder, interaction, elimination and selection modules as shown in Figure \ref{fig:diagram}. Among these, the main contribution of our work is the introduction of a module for elimination. Specifically, we introduce a module to (i) decide whether an option can be eliminated (ii) refine the passage representation to account for eliminated/un-eliminated options and (iii) repeat this process multiple times. In the remainder of this section, we describe the various components of our model.

\paragraph{Encoder Module:} We first compute vectorial representations of the question and options. We do so by using a bidirectional recurrent neural network which contains two Gated Recurrent Units (GRU) \cite{DBLP:journals/corr/ChungGCB14}, one which reads the given string (question or option) from left to right and the other which reads the string from right to left. For example, given the question $Q = [w_1^q, w_2^q,\ldots, w_N^q ]$, each GRU unit computes a hidden representation for each time-step (word) as:
\begin{align*}
\overrightarrow{h_i^q} = \overrightarrow{GRU_q} (\overrightarrow{h_{i-1}^q} , e(w_i^q)) \\
\overleftarrow{h_i^q} = \overleftarrow{GRU_q} (\overleftarrow{h_{i-1}^q} , e(w_i^q))
\end{align*}
where $e(w_i^q) \in \mathbb{R}^d$ is the $d$-dimensional embedding of the question word $w_i^q$. The final representation of each question word is a concatenation of the forward and backward representations (\textit{i.e.}, $h_i^q = [\overleftarrow{h_i^q}, \overrightarrow{h_i^q}]$). Similarly, we compute the bi-directional representations for each word in each of the $k$ options as $h_i^{z_k} = [\overleftarrow{h_i^{z_k}}, \overrightarrow{h_i^{z_k}}]$. Just to be clear, $h_i^{z_k}$ is the representation of the $i$-th word in the $k$-th option ($z_k$). We use separate GRU cells for the question and options, with the same GRU cell being used for all the $n$ options. Note that the encoder also computes a representation of each passage word as simply the word embedding of the passage word (\textit{i.e.}, $h_i^d = e(w_i^d)$). Later on in the interaction module we use a GRU cell to compute the interactions between the passage words.
\paragraph{Interaction Module:} Once the basic question and passage word representations have been computed, the idea is to allow them to interact so that the passage words' representations can be refined in the light of the question words' representations. This is similar to how humans first independently read the passage and the question and then read the passage multiple times, trying to focus on the portions which are relevant and ignoring portions that are irrelevant (\textit{e.g.}, portion marked in red in Figure \ref{fig:example}) to the question. To achieve this, we use the same multi-hop architecture for iteratively refining passage representations as proposed in Gated Attention Reader \cite{DBLP:conf/acl/DhingraLYCS17}.  At each hop $t$, we use the following set of equations to compute this refinement: 
\begin{align}
    	\nonumber \alpha^{t}_i &= \text{softmax}(Q^Td_i^{t})
\end{align}
where, $Q \in \mathbb{R}^{N \times l}$ is a matrix whose columns are $h_1^q, h_2^q, ..., h_N^q$ as computed by the encoder. $\alpha^{t}_{i} \in \mathbb{R}^N$ such that each element $j$ of $\alpha^{t}_{i}$ essentially computes the importance of the $j$-th question word for the $i$-th passage word during hop $t$. At the 0-th hop, $d^{0}_i = h_i^d = e(w_i^d) \in \mathbb{R}^l$ is simply the embedding of the  $i$-th passage word. The goal is to refine this embedding over each hop based on interactions with the question. Next, we compute,      
\begin{align}
\nonumber \tilde{q}^{t}_i &= Q\alpha^{t}_i
\end{align}
where $\tilde{q}^{t}_i \in \mathbb{R}^l$ computes the importance of each dimension of the current passage word representation and is then used as a gate to scale up or scale down different dimensions of the passage word representation. 
\begin{align}
        \nonumber \tilde{d}^{t}_i &= d^{t}_i \odot \tilde{q}^{t}_i
\end{align}
We now allow these refined passage word representations to interact with each other using a bi-directional recurrent neural network to compute $d^{(t+1)}_i$ for the next hop. 
 		\begin{align}
        	\nonumber \overrightarrow{d^{(t+1)}_{i}} &= \overrightarrow{\text{GRU}}_D^{(t+1)}(\overrightarrow{d^{(t+1)}_{i-1}},\tilde{d}^{(t)}_{i})\\
            \nonumber \overleftarrow{d^{(t+1)}_{i}} &= \overleftarrow{\text{GRU}}_D^{(t+1)}(\overleftarrow{d^{(t+1)}_{i-1}},\tilde{d}^{(t)}_{i})\\
            \nonumber
            d^{(t+1)}_{i} &= [\overleftarrow{d^{(t+1)}_{i}}, \overrightarrow{d^{(t+1)}_{i}}]
 		\end{align}
The above process is repeated for $T$ hops wherein each hop takes $d^{(t)}_i, Q$ as the input and computes a refined representation $\tilde{d}^{(t+1)}_i$. After $T$ hops, we obtain a fixed-length vector representation of the passage by combining the passage word representations using a weighted sum.
\begin{align}
    \nonumber m_{i} &= \text{softmax}(\tilde{d}^{(T)}_{i}W_{att}h^{q}_{N})\\ 
	\label{doc} x &= \sum_{i=1}^{M}m_{i}\tilde{d}^{(T)}_{i}
\end{align}
where $m_{i}$ computes the importance of each passage word and $x$ is a weighted sum of the passage representations.
\paragraph{Elimination Module:} The aim of the elimination module is to refine the passage representation so that it does not focus on portions which correspond to irrelevant options. To do so we first need to decide whether an option can be eliminated or not and then ensure that the passage representation gets modified accordingly. For the first part, we introduce an \textit{elimination} gate to enable a soft-elimination.
\begin{align}
				\nonumber e_i = \text{sigmoid}(W_ex + V_eh^{q} + U_eh^{z_i})
\end{align}
Note that this gate is computed separately for each option $i$. In particular, it depends on the final state of the bidirectional option GRU ($h^{z_i} = h^{z_i}_{J_i}$). It also depends on the final state of the bidirectional question GRU ($h^{q} = h^{q}_{N}$) and the refined passage representation ($x$) computed by the interaction module. $W_e, V_e, U_e$ are parameters which will be learned.

Based on the above soft-elimination, we want to now refine the passage representation. For this, we compute $x^{e}_{i}$ which is the component of the passage representation ($x$) orthogonal to the option representation ($h^{z_i}$) and $x^{r}_{i}$ which is the component of the passage representation along the option representation. 
\begin{align}
		   \nonumber r_i &= \frac{<x,h^{z_i}>h^{z_i}}{|x|^{2}}\\
           \label{Xe} x^{e}_{i} &= x -  r_{i} \\
           \label{Xk} x^{r}_{i} &= x -  x^{e}_{i}
    \end{align}
The \textit{elimination gate} then decides how much of $x^{e}_{i}$ and $x^{r}_{i}$ need to be retained.
\begin{align}
	\nonumber \tilde{x}_{i} &= e_{i}\odot x^{e}_{i} + (1 - e_{i})\odot x^{r}_{i}
\end{align}
If ${e}_{i} = 1$ (eliminate, \textit{e.g.}, portions corresponding to Option D in Figure \ref{fig:example}) then the passage representation will be made orthogonal to the option representation (akin to ignoring portions of the passage relevant to the option) and ${e}_{i} = 0$ (don't eliminate, \textit{e.g.}, portions marked in green, corresponding to Option A in Figure \ref{fig:example}) then the passage representation will be aligned with the option representation (akin to focusing on portions of the passage relevant to the option).

Note that in equations \eqref{Xe} and \eqref{Xk} we completely subtract the components along or orthogonal to the option representation. We wanted to give the model some flexibility to decide how much of this component to subtract. To do this we  introduce another gate, called the subtract gate,
\begin{align}
    \nonumber s_{i} &= \text{sigmoid}(W_{s}x+ V_{s}h^{q} + U_{s}h^{z_{i}})
\end{align}
where $W_s , V_s , U_s$ are parameters that need to be learned. We then replace the RHS of Equations \ref{Xe} and \ref{Xk} by $x - s_{i} \odot r_{i}$ and $x - s_{i} \odot x^e_{i}$ respectively.  Thus the components $r_i$ and $r^{\perp}_i$ used in Equation \eqref{Xe} and \eqref{Xk} are gated using $s_i$. One could argue that $e_i$ itself could encode this information but empirically we found that separating these two functionalities (elimination and subtraction) works better. 

For each of the $n$ options, we independently compute representations $\tilde{x}_{1}, \tilde{x}_{2}, ... ,\tilde{x}_{n}$. These are combined to obtain a single refined representation for the passage. 
\begin{align}
    \nonumber b_{i} &= v^{T}_{b}\text{tanh}(W_{b}\tilde{x}_{i} + U_{b}h^{z_{i}})\\
    \nonumber \beta_{i} &= \text{softmax}(b_{i})\\
	\label{tildeX} \tilde{x} &= \sum^{n}_{i=1}\beta_{i}\tilde{x}_{i}
\end{align}
Note that $\tilde{x}_{1}, \tilde{x}_{2}, ... ,\tilde{x}_{n}$ represent the $n$ option-specific passage representations and  $\beta_{i}$'s give us a way of combining these option specific representations into a single passage representation. We repeat the above process for $L$ hops wherein the $m$-th hop takes $\tilde{x}^{m-1}$, $h^q$ and $h^{z_i}$ as input and returns a refined $\tilde{x}^{m}$ computed using the above set of equations.
\paragraph{Selection Module} Finally, the selection module takes the refined passage representation $\tilde{x}^L$ after $L$ elimination hops and computes its bilinear similarity with each option representation.
\begin{align}
    \nonumber \text{score}(i) &= \tilde{x}^{L}W_{att}h^{z_{i}}
\end{align}
where $\tilde{x}^{L}$ and $h^{z_{i}}$ are vectors and $W_{att}$ is a matrix which needs to be learned. We select the option which gives the highest score as computed above. We train the model using the cross entropy loss by normalizing the above scores (using softmax) first to obtain a probability distribution.

\section{Experimental Setup}

\begin{figure}[!t]
\centering
\begin{subfigure}{.38\textwidth}
  \includegraphics[width=1\linewidth]{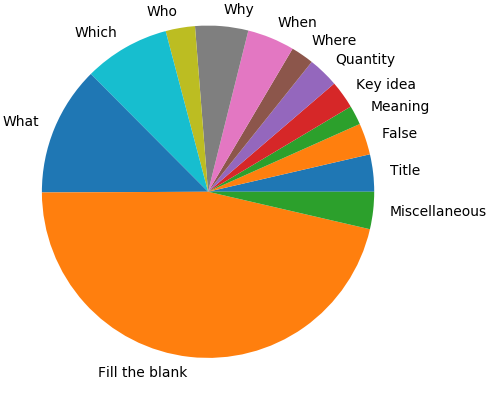}
\end{subfigure}%
\hfill
\begin{subfigure}{.38\textwidth}
  \includegraphics[width=1\linewidth]{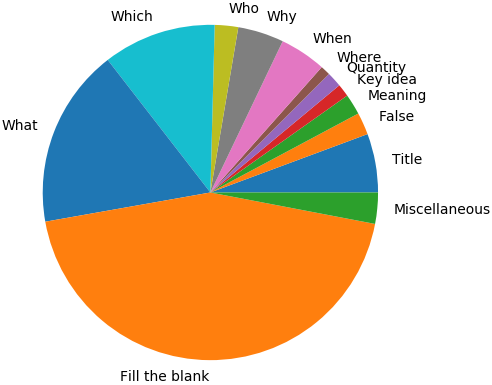}
\end{subfigure}
\caption{Distribution of different question types in the RACE-Mid (top) and RACE-High (bottom) portions of the dataset}
\label{fig:cat_dist1}
\end{figure}

In this section, we describe the dataset used for evaluation, the hyperparameters of our model, training procedure and state of the art models used for comparison.
\newline
\newline
\textbf{Dataset:} We evaluate our model on the RACE dataset which contains multiple choice questions collected from high school and middle school English examinations in China. The high school portion of the dataset (RACE-H) contains $62445$, $3451$ and $3498$ questions in train, validation, and test sets respectively. The middle school portion of the dataset (RACE-M) contains $18728$, $1021$ and $1045$ questions for train, validation, and test sets respectively. 
\newline
This dataset contains a wide variety of questions of varying degrees of complexity. For example, some questions ask for the most appropriate title for the passage which requires deep language understanding capabilities to comprehend the entire passage. There are some questions which ask for the meaning of a specific term or phrase in the context of the passage. Similarly, there are some questions which ask for the key idea in the passage. Finally, there are some standard Wh-type questions. Given this wide variety of questions, we wanted to see if there are specific types of questions for which an elimination module makes more sense. To do so, with the help of in-house annotators, we categorize the questions in the test dataset into the following $13$ categories using scripts with manually defined rules: (i) $6$ Wh-question types, (ii) questions asking for the title/meaning/key idea of the passage, (iii) questions asking whether the given statement is True/False, (iv) questions asking for a quantity (e.g., how much, how many) (v) fill-in-the-blanks questions. We were able to classify $91.26\%$ of questions in the test set into these $12$ categories and the remaining $8.74\%$ of questions were labeled as miscellaneous. 
The distribution of questions belonging to each of these categories in RACE-H and RACE-M are shown in Figure \ref{fig:cat_dist1}. 

\paragraph{Training Procedures:}
We try two different ways of training the model. In the first case, we train the parameters of all the modules (encoder, interaction, elimination, and selection) together. In the second case, we first remove the elimination module and train the parameters of the remaining modules. We then fix the parameters of the encoder and interaction module and train only the elimination and selection module. The idea was to first help the model understand the document better and later focus on elimination of options (in other words, ensure that the entire learning is focused on the elimination module). Of course, we also had to learn the parameters of the selection module from scratch because it now needs to work with the refined passage representations. Empirically, we find that this pre-training step does not improve over the performance obtained by end-to-end training. Hence, we report results only for the first case (\textit{i.e.}, end-to-end training).
\newline
\newline
\textbf{Hyperparameters:}
We restrict our vocabulary to the top 50K words appearing in the passage, question, and options in the dataset. We use the same vocabulary for the passage, question, and options. We use the same train, valid, test splits as provided by the authors. We tune all our models based on the accuracy achieved on the validation set. We initialize the word embeddings with $100$ dimensional Glove embeddings \cite{DBLP:conf/emnlp/PenningtonSM14}. We experiment with both fine-tuning and not fine-tuning these word embeddings. The hidden size for BiGRU is the same across the passage, question, and option and we consider the following sizes :$\{64, 128, 256\}$. We experiment with $\{1, 2, 3\}$ hops in the interaction module and $\{1,3,6\}$ passes in the elimination module. We add dropout at the input layer to the BiGRUs and experiment with dropout values of $\{0.2, 0.3, 0.5\}$. We try both Adam and SGD as the optimizer. For Adam, we set the learning rate to $10^{-3}$ and for SGD we try learning rates of $\{0.1, 0.3, 0.5\}$. In general, we find that Adam converges much faster. We train all our models for upto 50 epochs as we do not see any benefit of training beyond 50 epochs.
\newline
\newline
\textbf{Models Compared:} We compare our results with the current state of the art model on RACE dataset, namely, Gated Attention Reader \cite{DBLP:conf/acl/DhingraLYCS17}. This model was initially proposed for cloze-style RC and is, in fact, the current state of the art model for cloze-style RC. The authors of RACE dataset adapt this model for RC-MCQ by replacing the output layer with a layer which computes the bilinear similarity between the option and passage representations.

\begin{figure}[!b]
\begin{subfigure}{0.49\textwidth}
\includegraphics[width=1\textwidth]{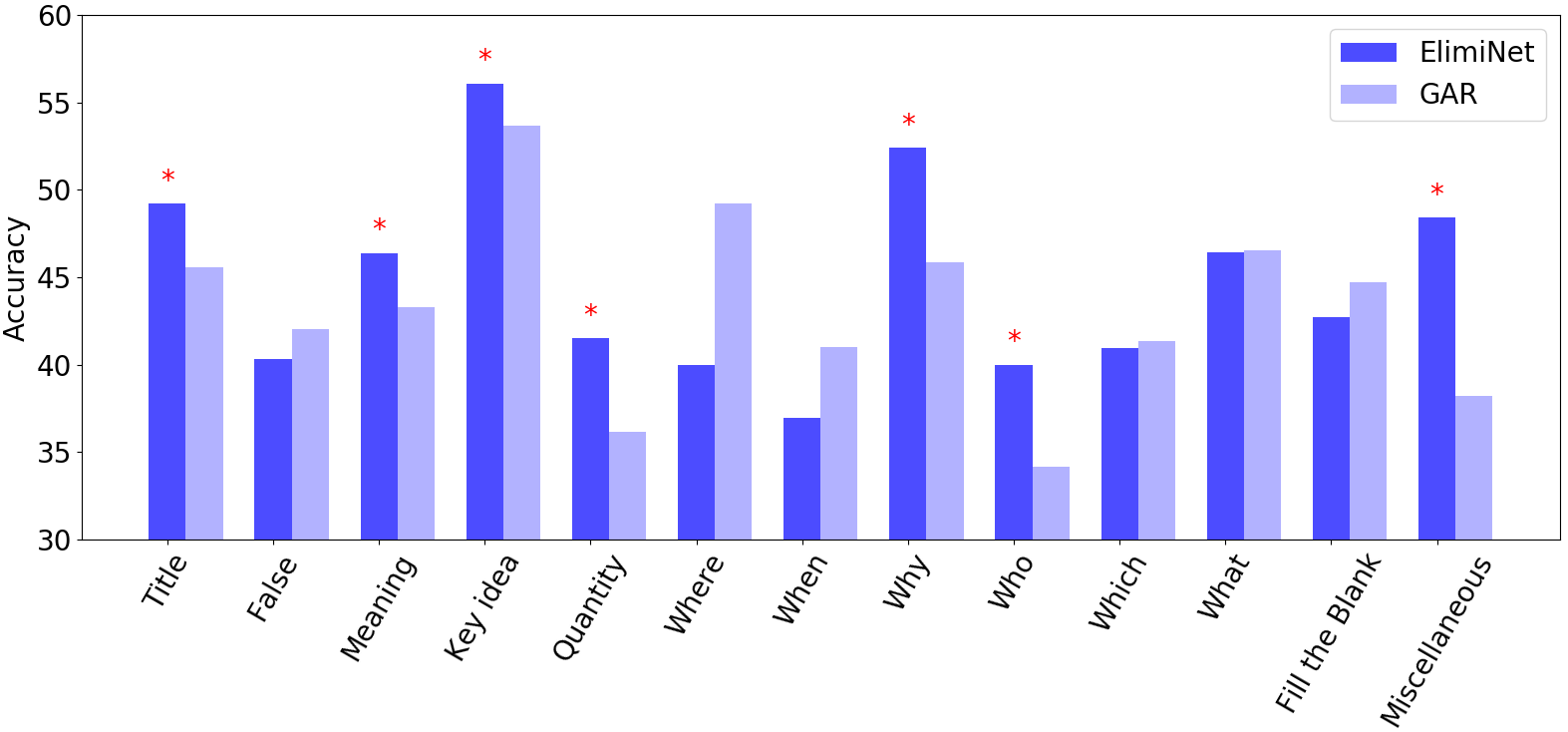}
\end{subfigure}
\begin{subfigure}{0.49\textwidth}
  \includegraphics[width=1\linewidth]{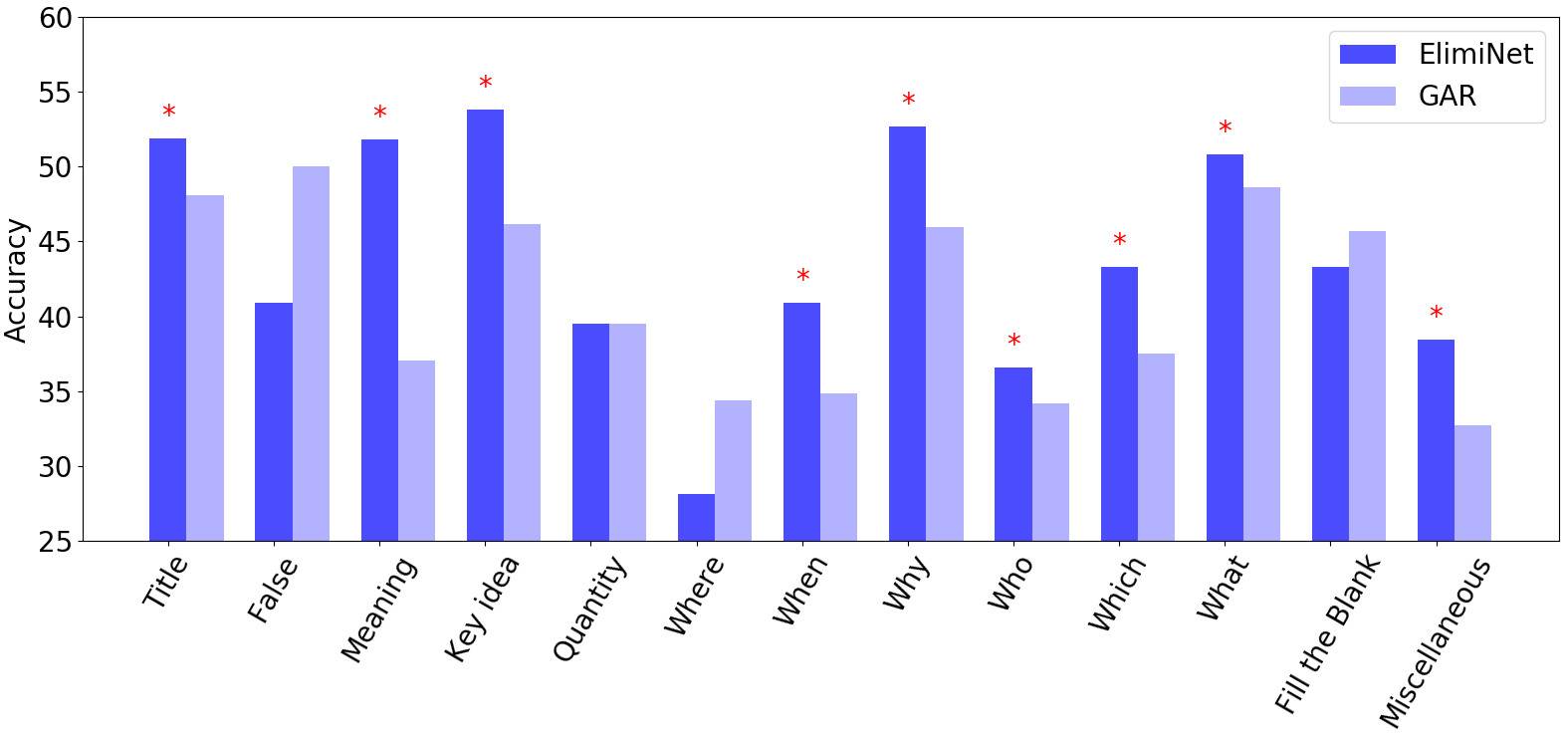}
\end{subfigure}
\begin{subfigure}{0.49\textwidth}
  \includegraphics[width=1\linewidth]{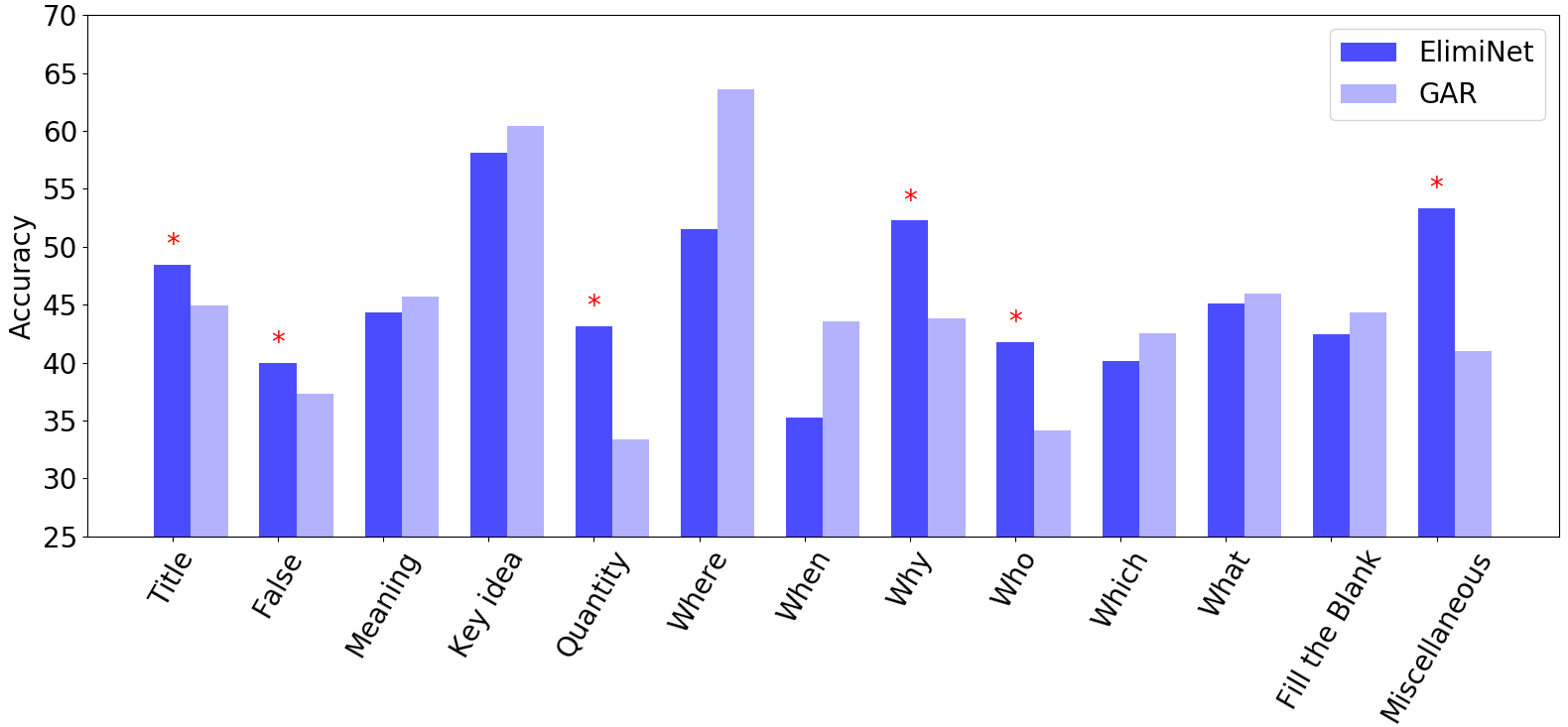}
\end{subfigure}
\caption{Performance of ElimiNet and Gated Attention Reader (GAR) on different question categories in RACE-Full (top), RACE-Mid (mid) and RACE-High (bottom). The categories in which our model outperforms GAR are marked with *.}
\label{perf_comp_mid}
\end{figure}

\section{Results and Discussions}

In this section, we discuss the results of our experiments. 

\subsection{Performance of Individual Models}

We compare the accuracy of different models on RACE-Mid (middle school), RACE-High (high school) and full RACE test-set comprising of both RACE-Mid and RACE-High. For each dataset, we compare the accuracy for each question type. These results are summarized in Figure \ref{perf_comp_mid}. We observe that on RACE-Mid ElimiNet performs better than Gated Attention Reader (GAR) on $9$ out of $13$ categories. Similarly, on RACE-High ElimiNet performs better than GAR on $6$ out of $13$ categories. Finally, on RACE-full, ElimiNet performs better than GAR on $7$ out of $13$ categories. Note that, overall on the entire test set (combining all question types) our model gives a slight improvement over GAR. The main reason for this is that the dataset is dominated by fill in the blank style questions and our model performs worse by only $2\%$ on such questions. However, since nearly $50\%$ of the questions in the dataset are fill in the blank style questions even a small drop in the performance on these questions, offsets the gains that we get on other question types.

\subsection{Ensemble of Different Models}

Since ElimiNet and GAR perform well on different question types we believe that taking an ensemble of these models should lead to an improvement in the overall performance. For a fair comparison, we also want to see the performance when we independently take an ensemble of $n$ GAR models and $n$ ElimiNet models. We refer to these as GAR-ensemble and ElimiNet-ensemble models. Each model in the ensemble is trained using a different hyperparameter setting and we use $n =6$ (we do not see any benefit of using $n > 6$). The results of these experiments are summarized in Table \ref{table:perfcomp}. ElimiNet-ensemble performs better than GAR-ensemble and the final ensemble gives the best results. We observe the ElimiNet-ensemble performs significantly better on RACE-Mid dataset than the GAR-ensemble and gives almost the same performance on the RACE-High dataset. Overall, by taking an ensemble of the two models we get an accuracy of $47.2\%$ which is $3.1\%$ (relative) better than GAR and  $1.3\%$ (relative) better than GAR-ensemble.

\subsection{Effect of Subtract Gate}

We wanted to see if the subtract gate enables the model to learn better (by performing partial orthogonalization/alignment). For this, we compared the accuracy with and without the subtract gate (we set the subtract gate to a vector of $1$s). We observed that the accuracy of our model drops from $44.33\%$ to $42.58\%$ and we outperformed the GAR model only in $3$ out of $13$ categories. This indicates that the flexibility offered by the subtract gate does help the model.

\begin{table}[!tbh]
\centering
\begin{tabular}{p{3.6cm}p{1.0cm}p{1.0cm}p{1.0cm}}
\hline
\textbf{Model} &  \textbf{RACE-Mid} & \textbf{RACE-High} & \textbf{RACE-Full} \\ \hline
SA Reader & 44.2 & 43.0 & 43.3 \\
GA Reader (GAR) &  43.7 & 44.2 & 44.1 \\
\textbf{ElimiNet }  & \textbf{44.4} & \textbf{44.5} & \textbf{44.5} \\
\cline{1-4}
GAR Ensemble & 45.7 & 46.2 & 45.9 \\
ElimiNet Ensemble &  \textbf{47.7} & 46.1 & 46.5 \\
GAR + ElimiNet (ensemble of above 2 ensembles) & 47.4 & \textbf{47.4} & \textbf{47.2} \\ 
\cline{1-4}
\end{tabular}
\caption{Performance of individual and ensemble models}
\label{table:perfcomp}
\end{table}

\begin{figure}[!t]
\centering
\begin{subfigure}{0.325\textwidth}
  \includegraphics[width=1\linewidth]{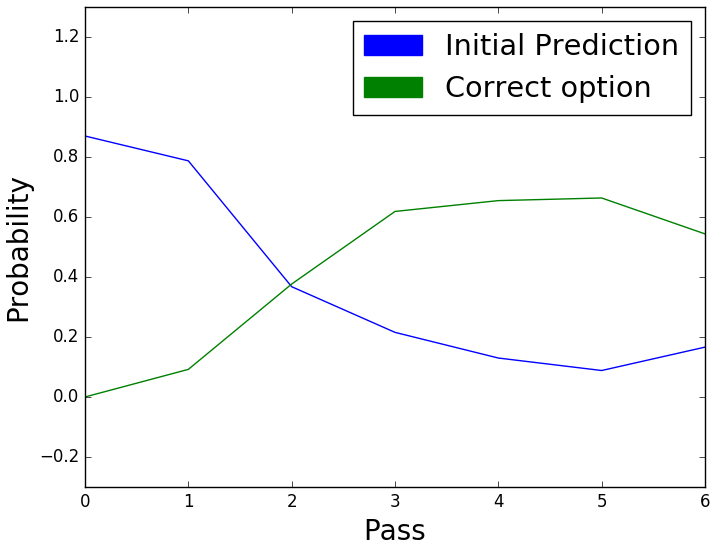}
\end{subfigure}%
\hfill
\begin{subfigure}{0.325\textwidth}
  \includegraphics[width=1\linewidth]{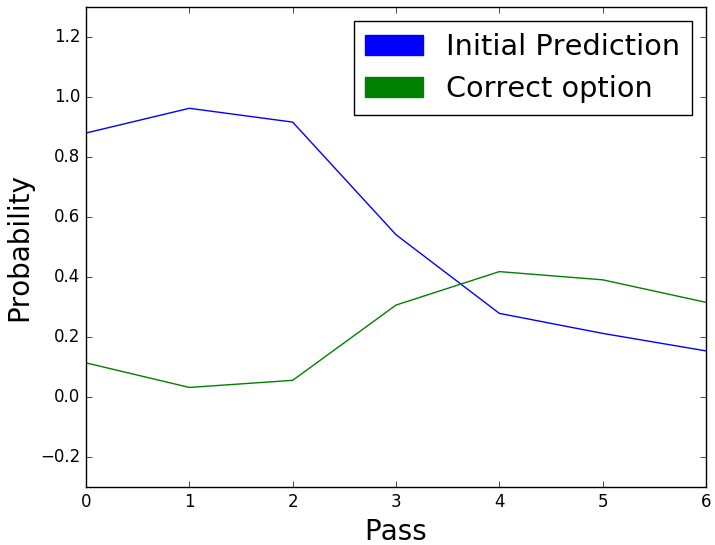}
\end{subfigure}

\caption{Change in the probability of correct option and incorrect option (initially predicted with highest score) over multiple passes of the \textit{elimination} module. The two figures correspond to two different examples from the test set.}
\label{fig:vis}
\end{figure}

\subsection{Visualizing Shift in Probability Scores}

If the elimination module is indeed learning to eliminate options and align/orthogonalize the passage representation w.r.t the uneliminated/eliminated options then we should see a shift in the probability scores as we do multiple passes of elimination. To visualize this, in Figure \ref{fig:vis}, we plot the probabilities of the correct option and the incorrect option with the highest probability before passing through $elimination$ module for two different test instances. We observe that as we do multiple passes of elimination, the probability mass shifts from the incorrect option (blue curve) to the correct option (green curve). This indicates that the elimination module is learning to align the passage representation with the correct option (hence, increasing its similarity) and moves it away from the incorrect option (hence, decreasing its similarity).

\section{Conclusion}

We focus on the task of Reading Comprehension with Multiple Choice Questions and propose a model which mimics how humans approach this task. Specifically, the model uses a combination of elimination and selection to arrive at the correct option. This is achieved by introducing an elimination module which takes a soft decision as to whether an option should be eliminated or not. It then modifies the passage representation to either align it with uneliminated options or orthogonalize it to eliminated options. The amount of orthogonalization or alignment is determined by two gating functions. This process is repeated multiple times to iteratively refine the passage representation. We evaluate our model on the recently released RACE dataset and show that it outperforms current state of the art models on $7$ out of $13$ question types. Finally, using an ensemble of  our elimination-selection approach with a state of the art selection approach, we get an improvement of $3.1\%$ over the best reported performance on RACE dataset. As future work, instead of soft elimination we would like to use reinforcement learning techniques to learn a policy for hard elimination.

\section*{Acknowledgments}

We thank Google for supporting Preksha Nema through their Google India Ph.D. Fellowship program. We thank our anonymous reviewer for suggesting the butterfly example which is mentioned in the introduction.

\bibliographystyle{named}
\bibliography{ijcai18}

\end{document}